\documentclass{article}

 \usepackage[main,final]{neurips_2025}

\usepackage[utf8]{inputenc} 
\usepackage[T1]{fontenc}    
\usepackage{hyperref}       
\usepackage{url}            
\usepackage{booktabs}       
\usepackage{amsfonts}       
\usepackage{nicefrac}       
\usepackage{microtype}      
\usepackage{xcolor}         
\usepackage{multirow}
\usepackage{makecell}

\usepackage{amsmath}
\usepackage{amssymb}
\usepackage{mathtools}
\usepackage{amsthm}
\usepackage{booktabs} 
\usepackage{float}
\usepackage{subcaption}
\usepackage{cuted}
\usepackage{capt-of}
\usepackage{amssymb} 
\usepackage[table]{xcolor}
\usepackage{authblk}

\title{Bridging the Indoor-Outdoor Gap: Vision-Centric Instruction-Guided Embodied Navigation}

\author[1*]{Yuxiang Zhao}
\author[1*†]{Yirong Yang}
\author[1‡]{Yanqing Zhu}
\author[1]{Yanfen Shen}
\author[1]{Chiyu Wang}
\author[1]{Zhining Gu}
\author[1]{Pei Shi}
\author[1]{Wei Guo}
\author[1]{Mu Xu}

\affil[1]{AMAP CV Lab, Alibaba Group}

\date{
  \textsuperscript{*}Equal contribution. \quad
  \textsuperscript{†}Work done during an internship at AMAP CV Lab. \quad
  \textsuperscript{‡}Project leader. \\
  Corresponding authors: \texttt{zhaoyao.zyx@alibaba-inc.com}, \texttt{xumu.xm@alibaba-inc.com}
}

\begin{document}
\begingroup
\raggedright
\setlength{\parindent}{0pt}%
\raisebox{-0.22\height}{\resizebox{!}{16pt}{\includegraphics{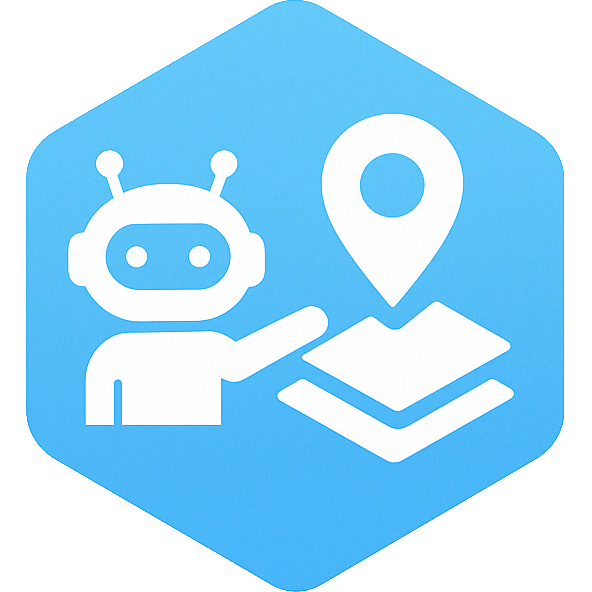}}}%
\hspace{1.2em}%
{\bfseries\large AMAP CV Lab}%
\hspace{2em}%
{\large Alibaba Group}%
\vspace{6pt}
\endgroup 

\maketitle

\noindent\begin{minipage}{\textwidth}
\vspace{-5em}
\centering
\footnotesize
\textsuperscript{*}Equal contribution. \quad
\textsuperscript{†}Work done during an internship at AMAP CV Lab. \quad
\textsuperscript{‡}Project leader. \quad
\texttt{zhaoyao.zyx@alibaba-inc.com}, \texttt{xumu.xm@alibaba-inc.com}
\end{minipage}

\begin{figure*}[htbp]
\centering
\includegraphics[width=1.0\linewidth]{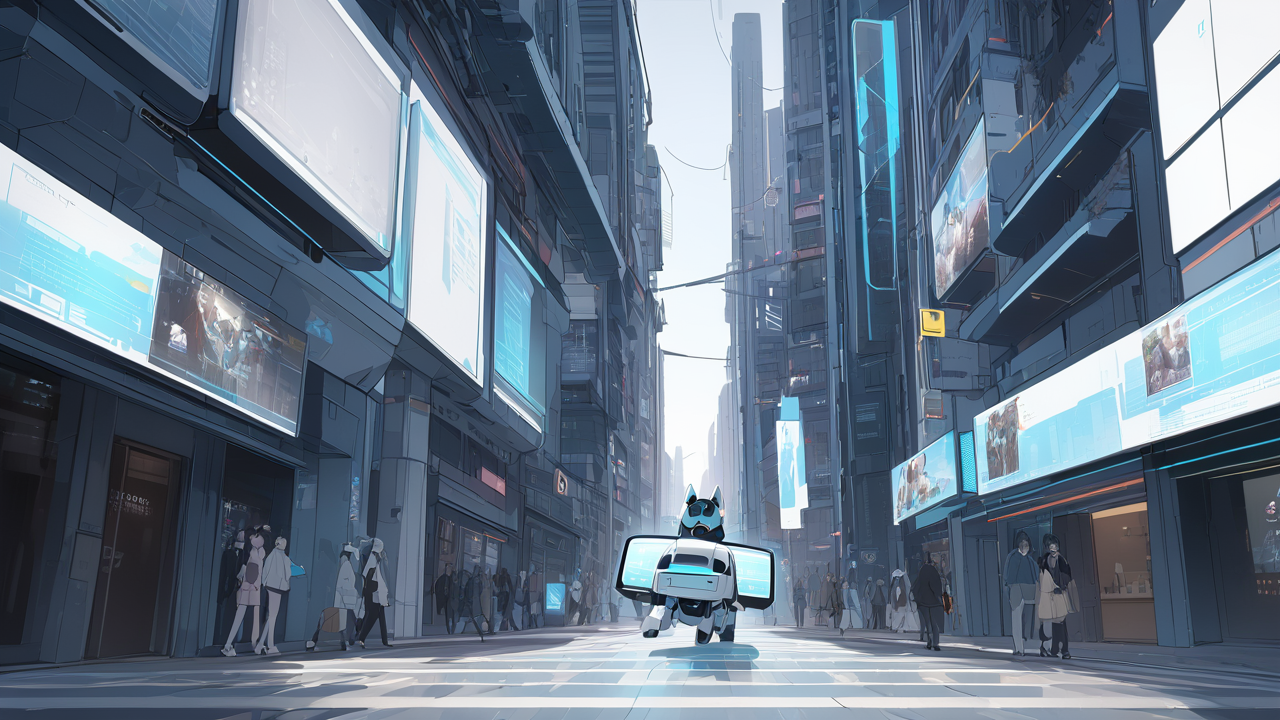}
\captionof{figure}{
Existing embodied navigation typically focuses exclusively on either indoor or outdoor scenes. However, embodied agents performing delivery tasks often need to transition seamlessly between these two environments. To bridge this gap, we propose a novel task BridgeNav that enables agents to navigate from outdoor to indoor and accurately enter buildings without relying on any additional priors.
}
\label{teaser}
\end{figure*}

\begin{abstract}
Embodied navigation holds significant promise for real-world applications such as last-mile delivery. However, most existing approaches are confined to either indoor or outdoor environments and rely heavily on strong assumptions—such as access to precise coordinate systems. While current outdoor methods can guide agents to the vicinity of a target using coarse-grained localization, they fail to enable fine-grained entry through specific building entrances, critically limiting their utility in practical deployment scenarios that require seamless outdoor-to-indoor transitions. To bridge this gap, we introduce a novel task: out-to-in prior-free instruction-driven embodied navigation. This formulation explicitly eliminates reliance on accurate external priors, requiring agents to navigate solely based on egocentric visual observations guided by instructions. To tackle this task, we propose a vision-centric embodied navigation framework that leverages image-based prompts to drive decision-making. Additionally, we present the first open-source dataset for this task, featuring a pipeline that integrates trajectory-conditioned video synthesis into the data generation process. Through extensive experiments, we demonstrate that our proposed method consistently outperforms state-of-the-art baselines across key metrics including success rate and path efficiency.
\end{abstract}

\section{Introduction}
Instruction-guided embodied navigation\cite{zhang2025embodied, zhang2024navid, zhai2023peanut, peng2025lovon, cheng2024navila, zhang2024uni} enables an embodied agent to reach a target location specified in instructions by generating a sequence of waypoints from visual observations. This capability supports applications such as robotic assistance, virtual reality, and smart home systems, attracting significant interest from academia and industry.

Previous research has largely focused on indoor, structured environments \cite{li2023layout, li2025regnav, lei2024instance}. In recent years, embodied navigation in outdoor scenes \cite{xu2025flame, chen2019touchdown, schumann2021generating} has garnered increasing attention. However, real-world outdoor navigation is significantly more challenging. Existing methods often rely on additional information, such as precise positional coordinates \cite{Citywalker, wang2024interactive} for point-to-point navigation, semantic maps \cite{chaplot2020object, yu2024trajectory} for embodied agents, or lengthy instructions that describe the agent’s surroundings in detail. Yet such auxiliary information is frequently inaccurate or difficult to obtain in practical deployment scenarios. On one hand, available location data typically refers to the target as a whole and lacks fine-grained details such as entrances and exits. At the same time, legal privacy protections restrict access to precise location and semantic maps. Consequently, existing approaches \cite{Habitat, Dd-ppo, wang2018look} that depend on positional coordinates for outdoor point-to-point navigation usually reach only the vicinity of the target and fail to enter the destination through its entrance to complete downstream tasks such as package pickup or delivery. On the other hand, requiring users to provide verbose textual instructions that meticulously describe both the surrounding environment and task procedures hinders real-world usability and deployment \cite{xu2025flame, schumann2021generating}. Therefore, while current outdoor embodied navigation pipelines that leverage auxiliary information can bring agents close to their destinations, there remains an urgent need for a solution that accomplishes the last mile of navigation without relying on any such extra information.

\textbf{\textit{To address the above challenges, we introduce a novel task: out-to-in prior-free instruction-driven embodied navigation, along with its corresponding solution, BridgeNav.}} The overall pipeline of our framework and its comparison with existing benchmarks are illustrated in Fig. \ref{teaser} and Table \ref{tab:dataset_comparison}, respectively. In this task, an embodied agent must navigate to a target destination using only its visual observations and simple textual instructions such as "go to Starbucks", without relying on any external prior information. This task represents the critical transition phase from outdoor to indoor navigation and constitutes a key component of the end-to-end embodied navigation pipeline, yet it remains underexplored. \textbf{\textit{In this work, our proposed BridgeNav is a novel image-information-driven, instruction-guided embodied navigation framework designed to fully unleash the model’s capacity to exploit visual information in the absence of any auxiliary cues.}} We observe that, as an embodied agent navigates from its starting position toward the target location, the visual content it should prioritize continuously evolves. Specifically, at the initial stage of navigation, the agent’s primary task is to determine whether the target is present in its current observation; during mid-course navigation, its focus shifts to interpreting signage to infer the general direction of the target location; and upon approaching the vicinity of the target, the agent must precisely locate entrances or exits to accurately reach the destination. To address this dynamic attention requirement, we employ a latent intention module that guides the agent to attend to the most relevant regions of the current visual observation. Furthermore, we introduce an optical-flow-guided dynamic perception module, enabling the agent to anticipate which parts of the visual observation will be predominantly affected by its own motion and thereby endowing it with an imaginative capability to foresee key changes in future observations.

Moreover, existing mainstream benchmarks \cite{R2R, HM3D, schumann2021generating, Matterport3D} are limited to lane-centric scenes for autonomous driving, indoor structured environments, or outdoor street view settings that lack natural language instructions and explicit destination goals. Scenarios in which embodied agents navigate outdoors to specific stores or landmarks guided by instructions and aimed at retrieving or delivering items are highly relevant to daily life yet remain underexplored. To address this gap, we introduce a large scale dataset BridgeNavDataset. \textbf{\textit{To the best of our knowledge, this is the first open-source dataset specifically designed for outdoor-to-indoor vision-and-language navigation. We propose a data production pipeline and innovatively integrate trajectory-conditioned video generation techniques.}} Our contributions are summarized as follows:
\begin{itemize}
\item We define a novel task: out-to-in prior-free instruction-driven embodied navigation, which enables agents to precisely reach target locations and bridges the gap between outdoor and indoor navigation.
\item We propose BridgeNav, a novel image-centric embodied navigation framework. By integrating a latent intention extraction module and an optical-flow-guided dynamic perception module, our agent fully leverages visual information, continuously “perceiving” and “imagining” the most critical regions within images.
\item To demonstrate the effectiveness and efficiency of BridgeNav, we evaluate it using a comprehensive set of metrics and compare it fairly with prior approaches under a unified experimental setting.
\item We release the first open-source dataset for out-to-in prior-free instruction-driven embodied navigation, created using an innovative trajectory-conditioned video generation method.
\end{itemize}

\begin{table*}[tbp]
\small
\centering
\setlength{\tabcolsep}{0.5mm}
\begin{tabular}{lccccccc}  
 \toprule
  \textbf{Benchmarks} & \textbf{Source} & \textbf{Scenes} & \textbf{Video} & \textbf{Spatial Span} & \textbf{Transit.} & \textbf{Consist.} & \textbf{Ann.} \\ \midrule
 VLN-CE\cite{VLN-CE} & MP3D & 90  & Simulator & R2R & N & N & A* \\
 OVON\cite{OVON} & HM3D & 181 & Simulator & R2R & N & N & A* \\
 GOAT-Bench\cite{Goat-bench} & HM3D & 181 & Simulator & R2R & N & N & A* \\
 \makecell[l]{SAGE-Bench\cite{SAGE-3D}} & InteriorGS & 1k & \makecell[c]{3DGS} & R2R & N & Y & A* \\ \midrule
 GNM\cite{Gnm} & GNM & -- & Sensor & Area-to-Area & N & Y & Sensor \\ \midrule
 Citywalker\cite{Citywalker} & HW & -- & Web & Area-to-Area & N & Y & VO \\
 UrbanNav\cite{urbannav} & HW & -- & Web & Area-to-Area & N & Y & VO \\
 \cmidrule{1-8}  
 Ours & Street View & 55k & Generation & Out-to-In & Y & Y & A* \\
\bottomrule
\end{tabular}
\caption{\textbf{Quantitative and structural comparison} between our proposed benchmark BridgeNav and representative indoor/outdoor benchmarks. InteriorGS \cite{InteriorGS2025}, HM3D \cite{HM3D}, and MP3D \cite{Matterport3D} are indoor-centric datasets; HW is sourced from the YouTube Human Walk videos. Spatial Span measures the environmental scale and semantic range. Video, Transit, Consist. and Ann. denote data source, inclusion of outdoor-to-indoor transitions, visual consistency between frames, and trajectory annotation method, respectively. VO is short for Visual Odometry. }
\label{tab:dataset_comparison}
\end{table*}

\section{Related Work}

\paragraph{Real-World Navigation.}
Utilizing real-world navigation data \cite{Gnm, ViNT, Lelan} for supervised learning is a prominent trajectory toward robust embodied agents. Some approaches aggregate multi-source robotic datasets \cite{Gnm, Open-x-embodiment, Lelan} to enhance cross-platform generalization and deployment efficiency. Others adopt generative paradigms, such as diffusion models \cite{Nomad, Deep-Active, Dreamernav}for trajectory synthesis or world models for future-state forecasting to guide planning.
Despite these advances, manual data collection is hindered by high costs and limited diversity. While leveraging web-scale data \cite{Omninav, Citywalker, NaviTrace, SocialNav} through automated pipelines and imitation learning offers a scalable alternative, the inherent noise in such data poses significant challenges. Even after filtering \cite{urbannav}, these models frequently internalize coarse correlations rather than precise, executable logic. However, such deviations lead to cumulative errors in intricate last-mile navigation. We develop the BridgeNav Dataset to leverage a trajectory-guided video generation framework to ensure precise terminal navigation through metric-aligned data.

\section{BridgeNav}
\subsection{Problem Formulation}
In this work, we address the problem of out-to-in prior-free instruction-driven embodied navigation. Within the overall embodied navigation pipeline, we assume that the agent can already reach the vicinity of the target location using a mature upstream point-to-point navigation module. The agent’s task is, upon arriving near the target area, to navigate from its current position to the exact goal location specified in the instruction and enter through an entrance or access point. At each time step $t$, the agent receives an RGB observation $o_{t}$ and an instruction $l$. The objective is to learn a policy $\pi(a_{t}|o_{t-k:t}, l)$ that maps a history of observations and positional information from the past few time steps to an action $a_{t}$ in the action space $\mathcal{A}$:

\begin{equation}
    a_{t} = \pi(a_{t}|o_{t-h:t}, l)
\end{equation}

This action is represented as a sequence of waypoints in Euclidean space. Typically, we set $h=10$ and predict actions for the next 5 steps simultaneously.

\subsection{Overview}
\begin{figure*}[tb]
  \centering
  \includegraphics[width=0.95\linewidth]{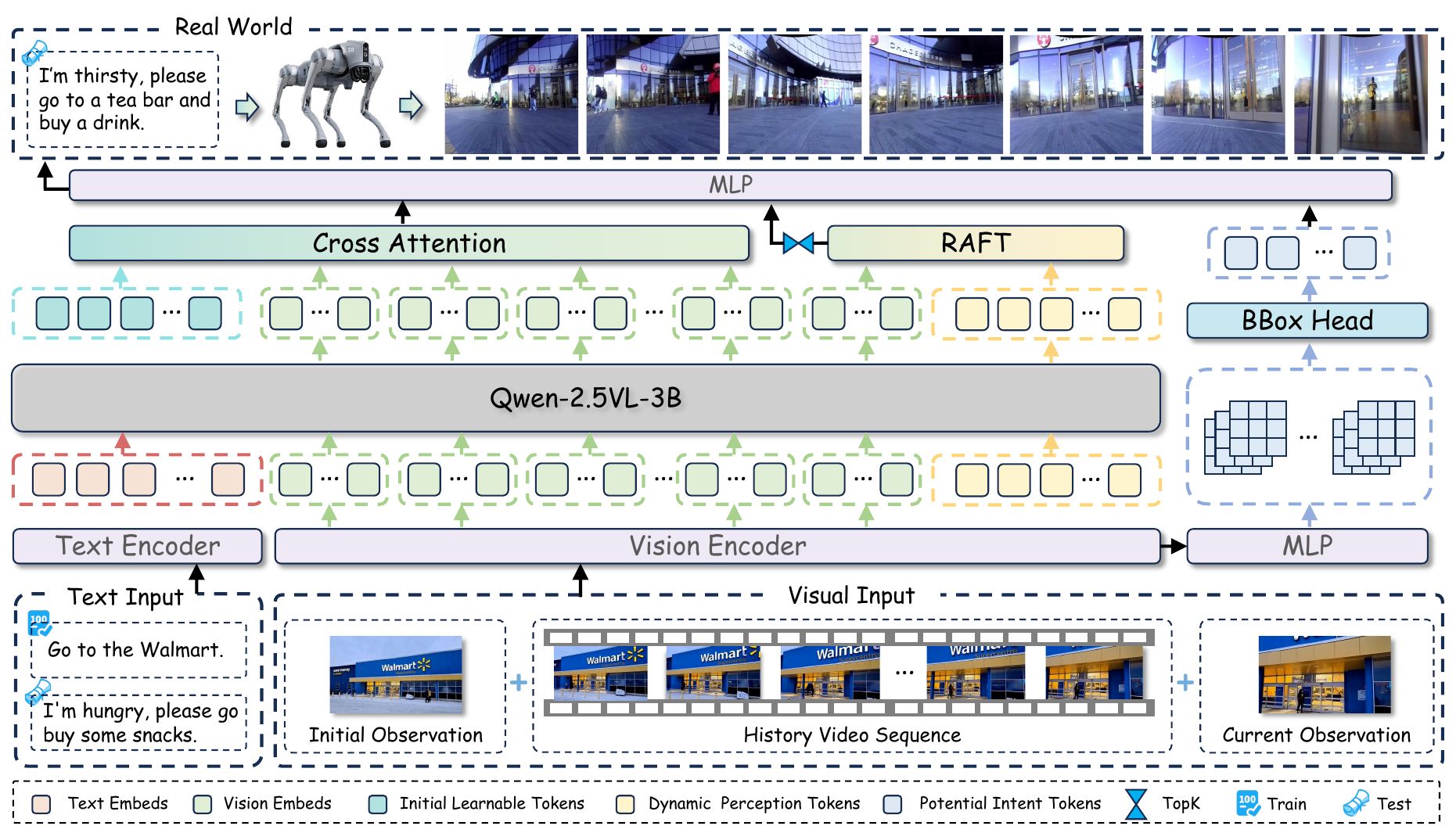}
  \caption{\textbf{Method overview.} Our proposed framework consists of four main components. (1) A multimodal large language model for vision–language understanding. (2) Cross-attention modules that enable interaction between initial learnable tokens (used for trajectory prediction) and the multimodal content representations. (3) A latent intention inference module that identifies salient regions in the current observation to guide attention. (4) An optical flow–guided dynamic perception module that establishes a mapping between the agent's navigation trajectory and salient changes in future visual observations.}
  \label{Fig.framework}
\end{figure*}

As illustrated in Fig. \ref{Fig.framework}, our proposed BridgeNav framework comprises four core modules. First, the natural language instruction $l$ and visual observation $o_{t}$ are processed by modality-specific encoders: the language instruction is encoded using a learnable embedding layer, and the image is encoded using a Vision Transformer. After encoding, the visual tokens are passed to the Latent Intention Inference module, which enables the agent to attend to task-relevant regions based on its current navigation phase; details are provided in Section \ref{sec3.3}. Concurrently, a set of learnable initial tokens is appended to these multimodal embeddings. These initial tokens are later used in the optical-flow-guided region reconstruction task, described in Section \ref{sec3.4}. Next, we employ a multimodal large language model based on Qwen2.5-VL-3B \cite{bai2025qwen2} to perform cross-modal integration and interaction, effectively transforming low-level perceptual signals into compact and coherent semantic representations. Finally, a decoder produces the predicted future navigation trajectory.
\subsection{Latent Intention Inference}
\label{sec3.3}
During the navigation of an embodied agent from a starting location to a target location, the distribution of observed image data changes significantly as the distance between the agent and the target location continuously decreases. \textbf{\textit{We argue that the most salient regions the agent should focus on within its visual observations also dynamically evolve throughout this process.}}

When the agent is far from the target location, its primary task is to assess the relevance between the observed image and the target described in the instruction, specifically to determine whether the target location appears within the current field of view. In practical deployment scenarios, if the target is not visible, the agent can seamlessly integrate with point-to-point navigation or exploration tasks.

When the agent is at a moderate distance from the target location, its main objective shifts to locating the signage associated with the target. At this stage, we contend that the agent should prioritize attending to the signage rather than the entrance itself, for two key reasons: firstly, the entrance occupies only a small region in the observed image due to the distance, making it challenging for the agent to learn its precise location, whereas the signage appears significantly larger in the image; secondly, the current observation often contains multiple candidate locations, and the visual differences among their signages are substantially more discriminative than those among their entrances.

When the agent is close to the target location, its primary goal becomes identifying and approaching the entrance to enter the target location. At this proximity, the entrance occupies a large portion of the image, and there is minimal interference from entrances of other candidate locations. Moreover, due to viewpoint constraints, the agent can no longer capture signage information reliably.

After obtaining the visual tokens encoded by a Vision Transformer, we pass them through multiple stacked Transformer blocks to extract features and then use a regression head to perform the grounding task. The ground-truth bounding boxes used for supervision were annotated during dataset construction using strategies specifically adapted to the characteristics of the image data distribution.

\subsection{Optical Flow-Guided Dynamic Perception}
\label{sec3.4}
Through the latent intention inference, we enable the agent to focus on the most salient regions in its current visual observation. Recent studies in Vision-Language-Action (VLA) \cite{Reconvla, Cosmos_Policy} learning have shown that endowing agents with the ability to "imagine" future states can significantly improve task performance. Motivated by this, we explore how such future imagination capabilities can be leveraged in Vision-and-Language Navigation (VLN).

In contrast to existing approaches, our method encourages the agent to imagine which regions of its visual observation will be most affected by its upcoming movement, thereby learning a mapping between the agent’s motion and the resulting salient changes in its visual input.

Specifically, during training, we employ a lightweight optical flow estimation method RAFT\cite{teed2020raft}, to compute the optical flow field $F\in \mathbb{R} ^{B\times 2\times H\times W}$ between the current observation $o_{t}$ and the next frame $o_{t+1}$. For each pixel location $(i,j)$ in batch $b$, the corresponding optical flow vector is given by 
\begin{equation}
F_{b,:,i,j} =\begin{bmatrix}
u_{b,i,j}  
\\
v_{b,i,j} 
\end{bmatrix}
\end{equation}
For each pixel location $(i, j)$, we compute the L2 norm of its optical flow vector:
\begin{equation}
M_{b,i,j}=\left \| F_{b,:,i,j}  \right \| _{2} =\sqrt{u_{b,i,j}^{2}+v_{b,i,j}^{2}  } 
\end{equation}
where $M=\left \| F \right \| _{2} \in \mathbb{R} ^{B\times 1\times H\times W} $.

Rather than reconstructing the entire future image, our goal is to help the agent perceive salient changes induced by its own motion and thereby learn the correspondence between ego-motion and visual dynamics. To this end, we select the top-$k$ pixel locations with the largest flow magnitudes from $F$ to generate a binary mask $Z\in {\left \{ 0,1 \right \} }^{B\times 1\times H\times W} $, and only reconstruct the corresponding masked regions.
\begin{equation}
Z_{b,i,j}=\begin{cases}
1, if (i,j)\in TopK_{b} 
 \\
0, otherwise
\end{cases}
\end{equation}
where $K_{b}$ denotes the set of spatial locations in the next-frame image with the largest optical flow magnitudes.
In our experiments, $k$ is set to 10\% of the total number of image pixels, though this value can be adjusted based on the specific task requirements.

After fusing the learnable initial tokens with multimodal embeddings in Qwen2.5-VL-3B for cross-modal aggregation, we extract these tokens and feed them into a salient-region decoder. This decoder consists of a linear projection layer followed by several transposed convolutional layers for upsampling, ultimately producing the agent’s “imagined” future image $x_{t+1}$.
\subsection{Training Strategy}

Our proposed BridgeNav is trained in a two-stage manner by minimizing the loss function $\mathcal{L}$. In the first stage, we primarily train the model’s potential intention inference capability. In the second stage, we freeze the potential intention inference branch and mainly train the model’s navigation capability, dynamic perception ability, and vision-instruction alignment. Specifically, navigation capability is supervised using trajectory waypoints, and dynamic perception is supervised by pixels within the core region exhibiting the largest optical flow magnitude, which corresponds to the area most affected by the agent’s motion. Note that the decoding module is only required during training to generate estimates of future core regions; during inference, the dynamic perception tokens can directly assist trajectory navigation without decoding. Additionally, we expect the model to return a vision-instruction alignment flag when the target location is not present in the current field of view, ensuring compatibility with downstream modules in embodied systems, such as triggering point-to-point navigation or exploration tasks. Furthermore, we artificially construct a set of negative samples by randomly swapping instructions and observation images across data samples, and treat this as a binary classification task to train the model to compute vision-language alignment based on the current instruction and observation. The overall loss function $\mathcal{L}$ is defined as follows:
\begin{equation}
    \mathcal{L}=\begin{cases}
    \mathcal{L}_{bbox} & \text{ if } stage1 \\
    \mathcal{L}_{wpts}+\mathcal{L}_{recon}+\mathcal{L}_{flag} & \text{ if } stage2
\end{cases}
\end{equation}
where $\mathcal{L}_{\text{bbox}}$ is the loss supervising the predicted latent intention bounding box, $\mathcal{L}_{\text{wps}}$ is the L1 distance between the predicted waypoints and the ground-truth waypoints, $\mathcal{L}_{\text{recon}}$ is the L1 distance between the predicted salient regions and the ground-truth counterparts, and $\mathcal{L}_{\text{flag}}$ is the BCE loss used to supervise the vision–instruction matching prediction.

\begin{figure*}[tb]
  \centering
  \includegraphics[width=0.93\linewidth]{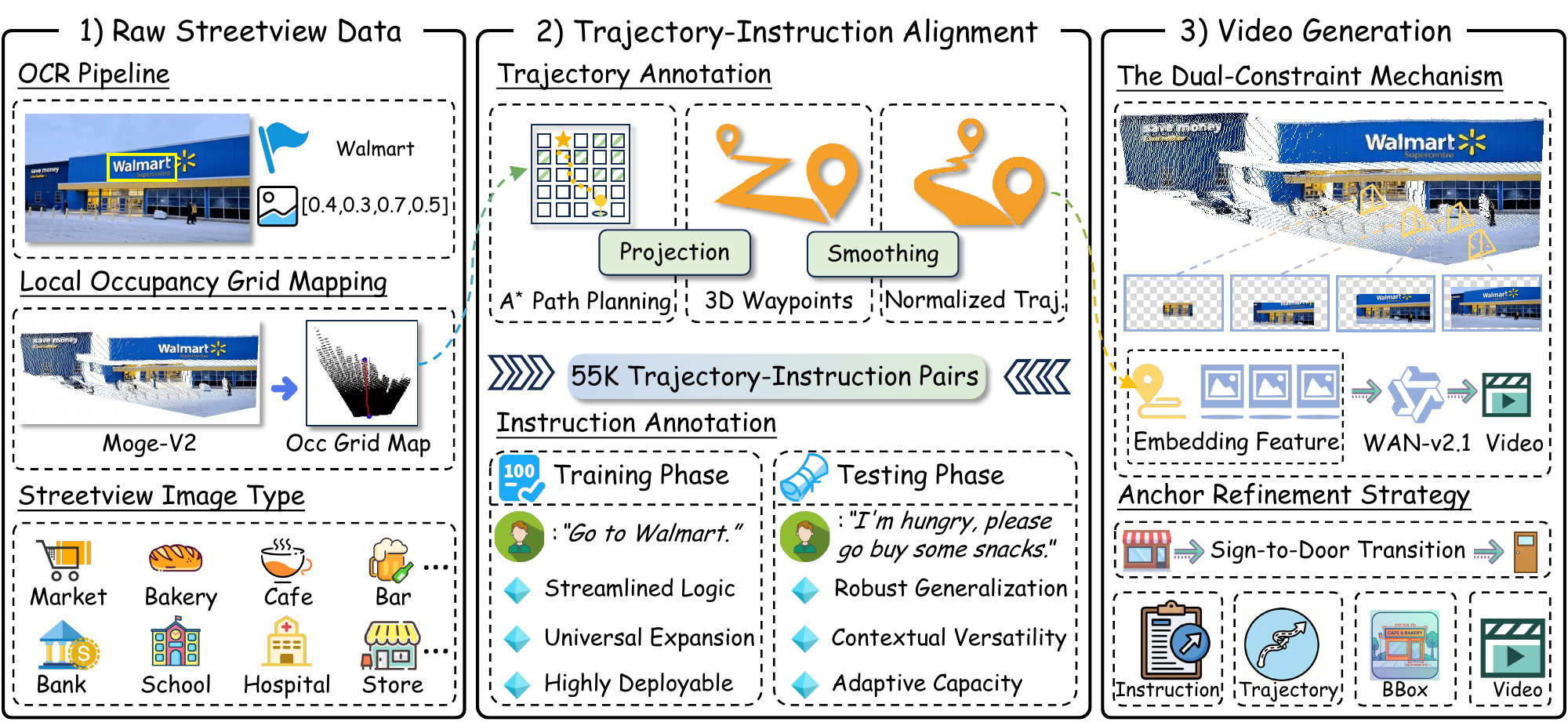}
  \caption{\textbf{Dataset construction overview.} Our proposed data generation pipeline consists of three main components. (1) Target location acquisition and occupancy map construction. (2) Trajectory and instruction  annotation. (3) Video synthesis under constraints from the initial frame and annotated trajectories.}
  \label{Fig.data_ppl}
\end{figure*}

\section{BridgeNavDataset}
\paragraph{Raw Streetview Data.}
\label{Raw Streetview Data}
To tackle the complexities inherent in the final stage of outdoor navigation, we curated an extensive dataset comprising 55K high-resolution egocentric images of urban street environments. These views capture a vast array of micro-semantic details, including house numbers, building entrances, and specific signage, which serve as detailed environmental anchors. Such visual cues effectively compensate for the limitations of traditional Global Positioning Systems or lane-based high definition maps in characterizing the complexities of terminal navigation scenes. Motivated by the accessibility of these data and their inherent alignment with navigation objectives, we leverage this rich visual information to establish a direct mapping from pixels to paths, thereby facilitating the training of agents for autonomous navigation without reliance on external priors. With the objective of extracting a unique navigation target from the visual pixels of each individual streetview image, we employ an automated Optical Character Recognition pipeline for semantic extraction and label denoising, as illustrated in Fig. \ref{Fig.data_ppl}. The target bounding boxes of identified street-level shop signs serve as the primary guidance to direct the agent toward designated destinations.

\paragraph{Trajectory and Instruction Annotation.}
Aiming for the extraction of position information, we perform trajectory annotation on the constructed local agent-target occupancy grid (Local occupancy grid mapping as detailed in Appendix \ref{Dataset Details}). First, we utilize the A star path planning algorithm \cite{3D-Mem} to compute an optimal obstacle-free path from the current agent position to the target, which serves as the 2D trajectory position. Subsequently, mapping the 2D grid path back into the 3D coordinate system is achieved through an inverse projection process, which recovers the depth and spatial height information for each waypoint. Building upon this, we derive the heading angles and poses of the agent at each stage through the calculation of direction vectors between waypoints. The resulting trajectories are characterized by the representation $[x, y, z, \text{yaw}]$. 
To guarantee physical feasibility, a standardized preprocessing workflow is applied, including trajectory origin normalization, sequence length resampling and path smoothing optimization. 

We introduce a differentiated instruction labeling strategy designed to account for sparse semantic inputs and improve zero-shot generalization across unseen scenes.
During the training phase, we employ standardized, concise instructions for focusing on core target semantics (such as "Go to Starbucks."), which establishes a direct mapping between high-level goals and spatial trajectories. In contrast, we introduce fuzzy natural language instructions during the testing phase (such as "Go to Starbucks and help me get a cup of coffee.") for the evaluation of the adaptive capacity of the model. Our instruction annotation strategy ensures that the navigation agent remains effective across diverse and unseen urban scenarios by reducing over-reliance on linguistic cues and emphasizing the understanding of spatial geometric relationships. The annotation pipeline ultimately yields 55k outdoor streetview navigation trajectory-instruction pairs.

\paragraph{Trajectory Guided Video Generation.}
Addressing the scarcity and high cost of annotated street-view data, we propose a trajectory-guided video generation framework utilizing the Wan2.1-I2V Diffusion Transformer. First, we utilize global point cloud features derived from MoGe-v2 predictions; specifically, each discrete point on the planned trajectory is treated as a virtual camera pose with independent position and orientation. By incorporating camera intrinsic and extrinsic parameters, the global point cloud is reprojected onto the 2D image plane corresponding to each pose, yielding a sequence of images that reflect the path's geometric structure. This reprojected sequence realistically simulates visual occlusions and field-of-view (FOV) variations during agent movement, serving as the primary geometric constraint. Second, we introduce plucker embedding to encode the virtual camera pose sequence, establishing a secondary constraint that ensures physical consistency between the generated video and the planned trajectory under perspective transformations. (The qualitative results of the video generation are shown in Appendix Fig. \ref{fig:show_case_4x1}.) This pipeline yielded a large-scale dataset of \textbf{55k} high-quality outdoor navigation trajectory-instruction pairs, encompassing \textbf{100+ hours} of video and over \textbf{10 million} images. Each video is rendered at $1440 \times 1080$ resolution and 23 fps, providing a robust foundation for subsequent model training.

\paragraph{Target Anchor Refinement.}
As the agent approaches the destination within the generated video sequences, shop signs serving as distant guidance landmarks gradually exit the frame due to FOV constraints. To maintain anchor continuity as the agent nears the target, we introduce an anchor refinement strategy. Specifically, we sampled 20k images from the generated video data and shifted the navigation target's bounding box from the shop sign to the shop door (entrance). By manually annotating the bounding box in these images, we provide a continuous navigation goal supervision signal for the first stage of model training.

\section{Experiments}
\subsection{Evaluation Metrics}
\paragraph{Task Success Rate.}  
This task primarily addresses the transitional phase between outdoor and indoor navigation, enabling embodied agents to enter target buildings through entrances or exits. In real-world scenarios, doors are often only partially open or obstructed by decorative elements on either side. Consequently, adopting a success threshold based solely on agent step size, as in prior work, is insufficient. Moreover, considering the physical width of the embodied agent, we report success rates within 0.1\,m, 0.2\,m, and 0.3\,m radii around the target location.
\paragraph{Navigation Efficiency.}In our dataset annotation, all ground-truth trajectories are generated using the A* path planning algorithm. To assess navigation efficiency, we report the deviation between the agent’s inferred trajectory and the ground-truth trajectory, as this discrepancy reflects how closely the agent approximates an optimal path.

\begin{table*}[htbp]
\small
\centering
\setlength{\tabcolsep}{0.4mm}
\begin{tabular}{l|ccc|ccc}
 \toprule
\multicolumn{1}{r|}{\textbf{Metrics}} & \multicolumn{3}{c|}{\textbf{Task Success Rate}} & \multicolumn{3}{c}{\textbf{Navigation Efficiency}} \\ 
  \textbf{Models}  & \textbf{SR (0.1m)} $\uparrow$ & \textbf{SR (0.2m)} $\uparrow$ &  \textbf{SR (0.3m)} $\uparrow$ & \textbf{TR (mean)} $\downarrow$ & \textbf{TR (best)} $\downarrow$ & \textbf{TR (worst)} $\downarrow$ \\ \midrule
NoMaD (ICRA'24) & 4.13 & 15.07 & 29.20 &  31.35 & 5.45 & 85.91 \\
Citywalker (CVPR'25) & 13.79 & 41.02 &  65.96 & 15.58 & 0.76 & 56.47 \\
OmniNav (ICLR'26) & 18.78 & 46.99 & 72.39 & 14.16 & 0.99 & 53.79 \\
\midrule
Ours & \textbf{33.82}\textcolor{red}{$_{80.1\%\uparrow}$} & \textbf{70.11}\textcolor{red}{$_{49.2\%\uparrow}$}  & \textbf{89.55} \textcolor{red}{$_{23.7\%\uparrow}$} &\textbf{9.77}\textcolor{red}{$_{31.0\%\downarrow}$} & \textbf{0.33}\textcolor{red}{$_{66.7\%\downarrow}$} & \textbf{51.38}\textcolor{red}{$_{4.5\%\downarrow}$} \\
\bottomrule
\end{tabular}
\caption{\textbf{Quantitative comparison} between our method and very recent vision-language navigation approaches, evaluating both task success rate and navigation efficiency.}
\label{tab:table2}
\end{table*}

\begin{table}[htbp]
\small
\centering
\setlength{\tabcolsep}{0.4mm}
\begin{tabular}{l|ccccc}
 \toprule
\textbf{Models} & \textbf{SR(0.1m)} $\uparrow$ & \textbf{SR(0.3m)} $\uparrow$ & \textbf{TR(mean)} $\downarrow$ \\ 
 \midrule
w/o learnable tokens & 5.64 & 42.89 & 22.39 \\
w/o intention inference & 15.24 & 73.54 & 14.05 \\
w/o dynamic perception & 29.20 & 86.55 & 10.69 \\
\midrule
Ours & 33.82 & 89.55 & 9.77 \\
\bottomrule
\end{tabular}
\caption{\textbf{Ablation study} of the main components in our framework.}
\label{tab:table3}
\end{table}


\begin{figure*}[tbp]
    \centering
    \includegraphics[width=0.328\linewidth]{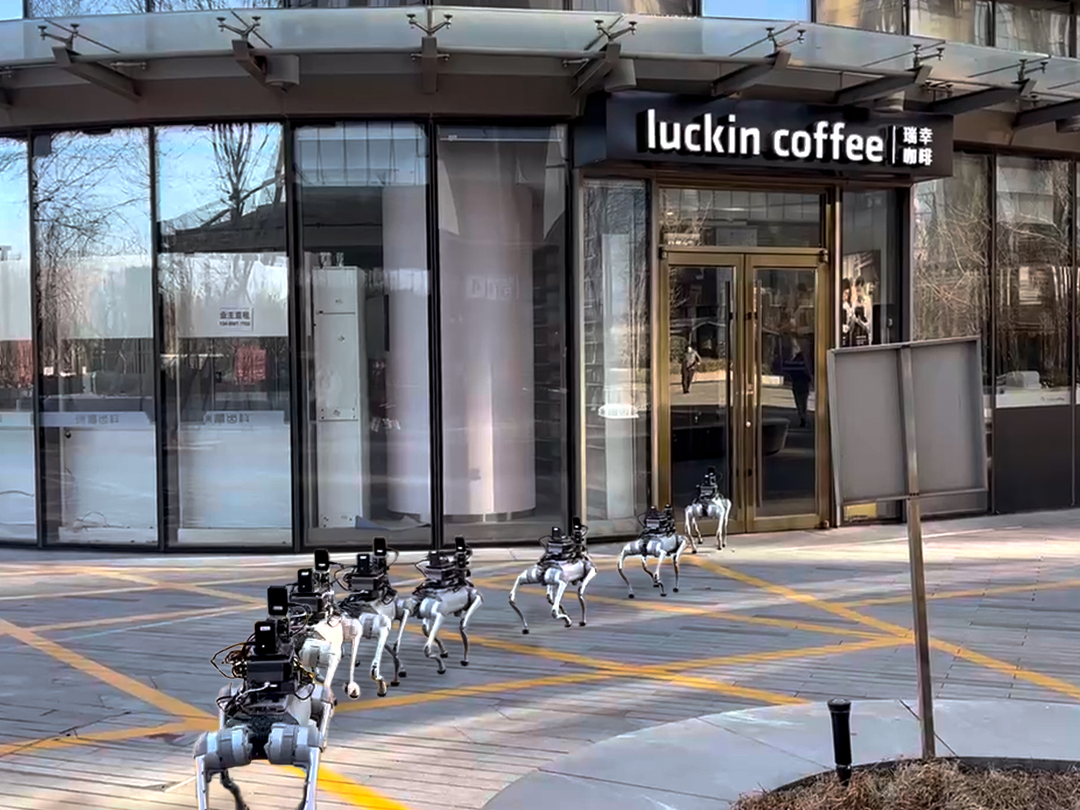}%
    \hfill
    \includegraphics[width=0.328\linewidth]{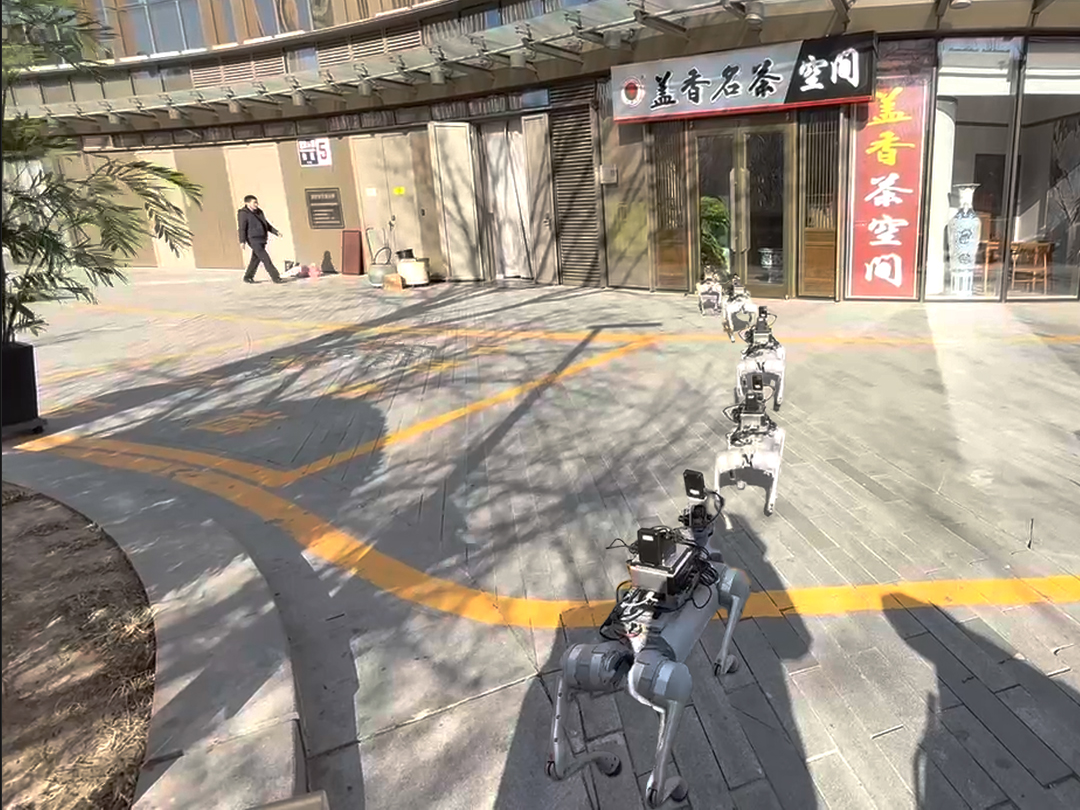}%
    \hfill
    \includegraphics[width=0.328\linewidth]{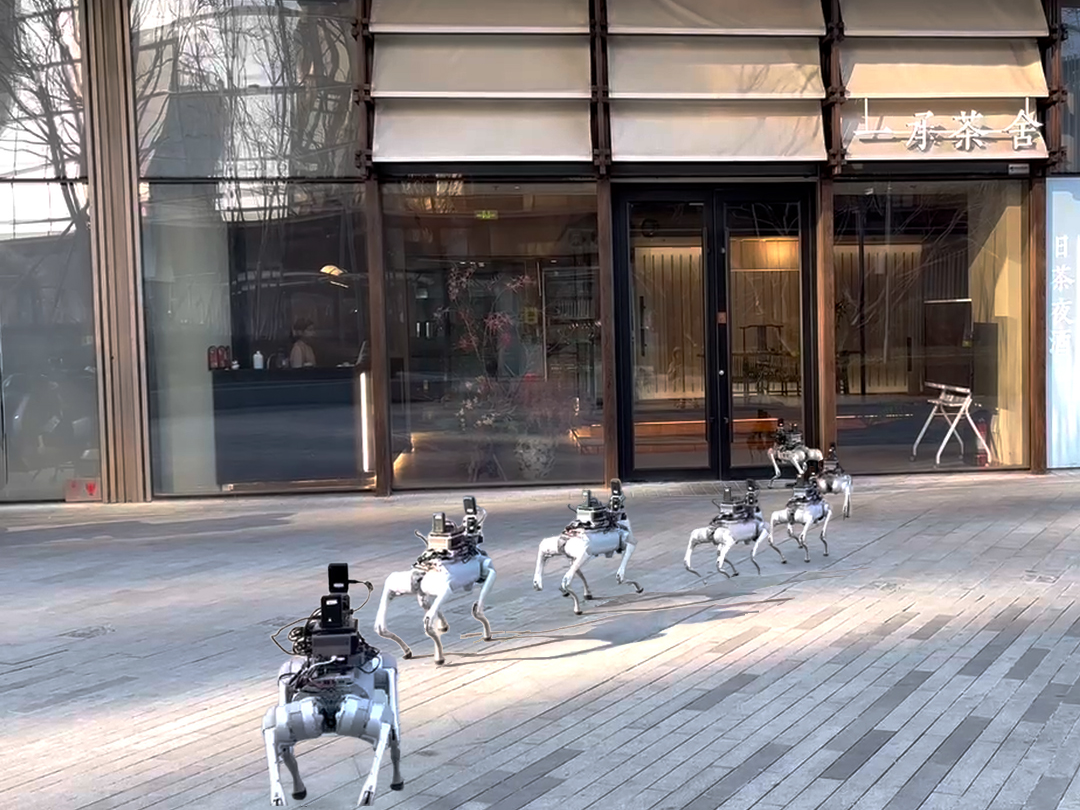}%
    \vspace{0.2em}
    \includegraphics[width=0.328\linewidth]{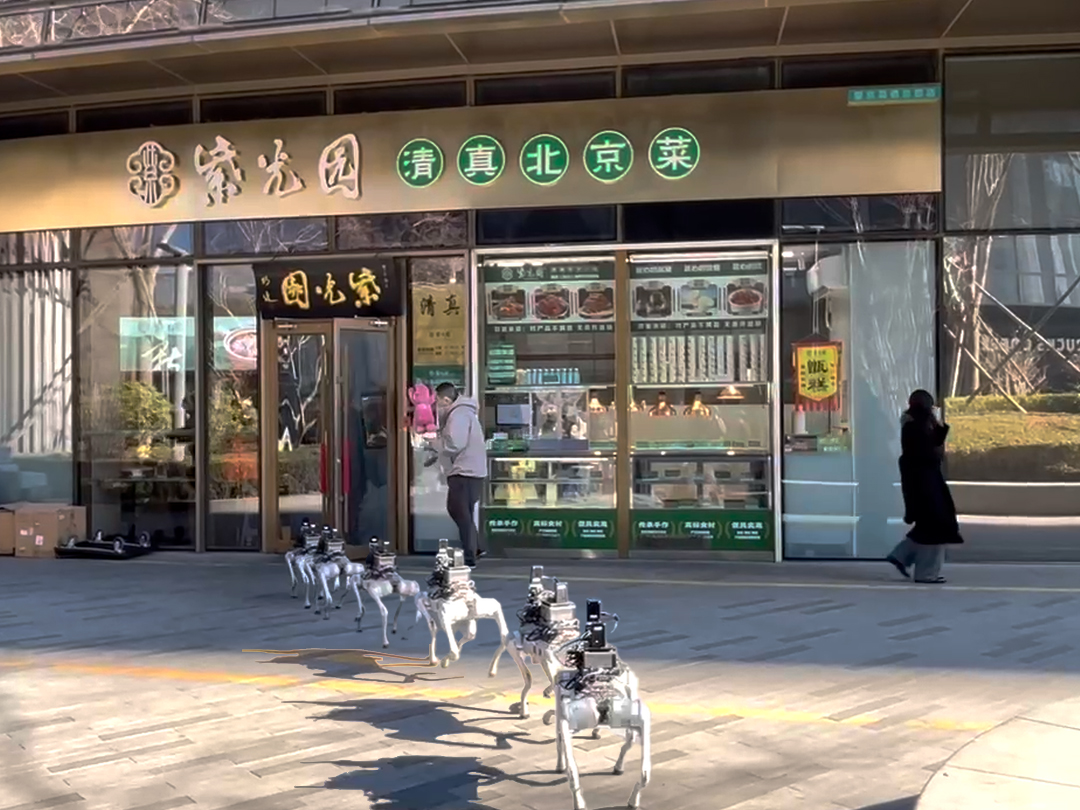}%
    \hfill
    \includegraphics[width=0.328\linewidth]{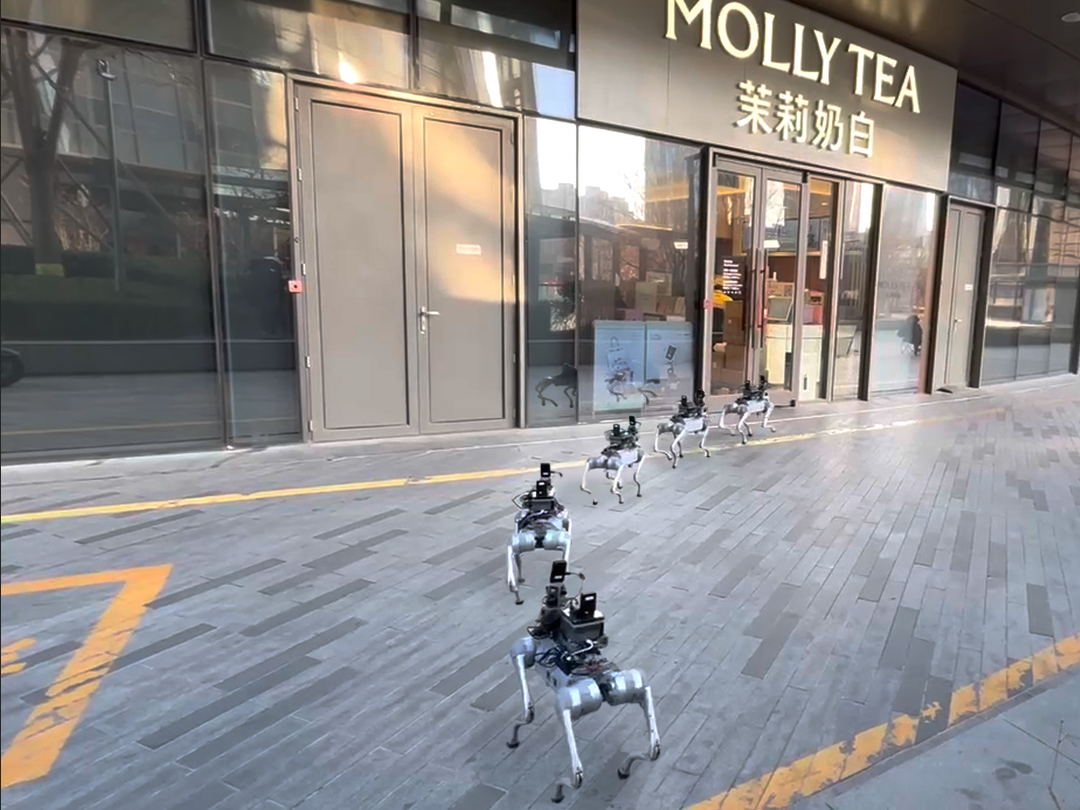}%
    \hfill
    \includegraphics[width=0.328\linewidth]{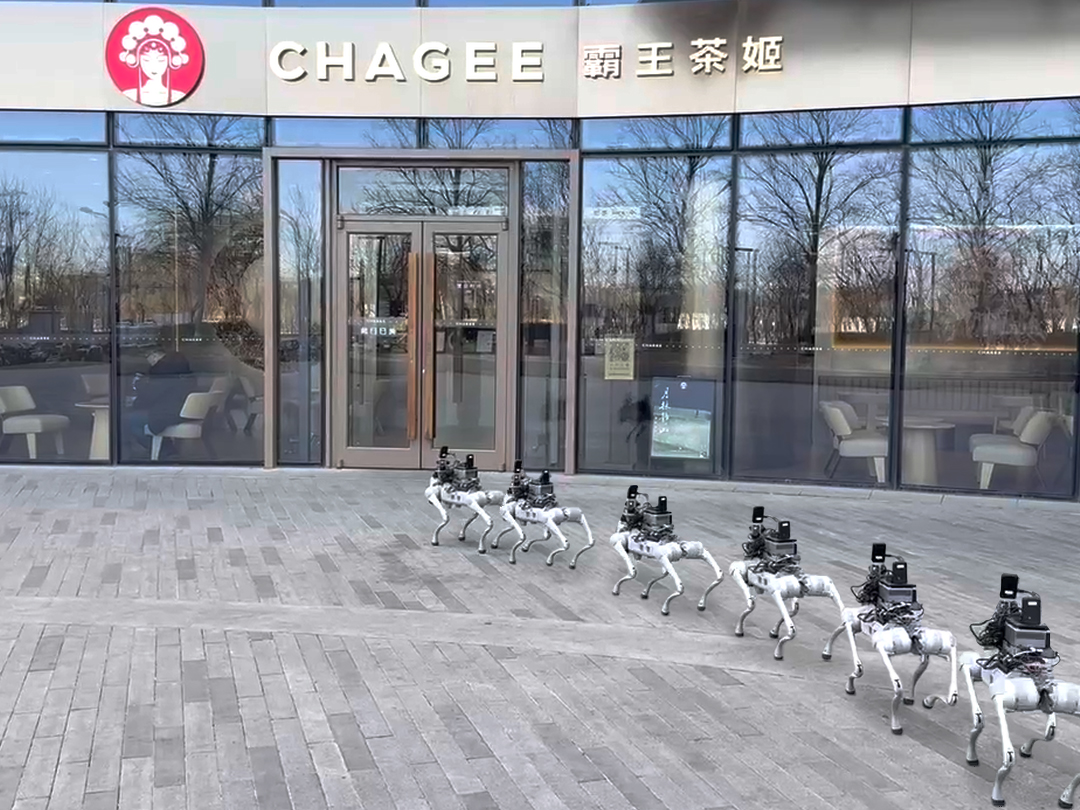}%
    \caption{\textbf{Qualitative results from real-world deployment.} Best viewed in color and zoomed in for more details.}
    \label{fig4}
\end{figure*}

\begin{figure}[htbp]
  \centering
   \includegraphics[width=0.32\textwidth]{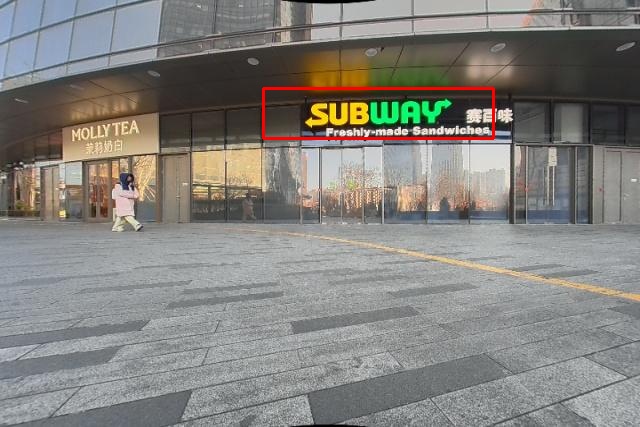}
   \includegraphics[width=0.32\textwidth]{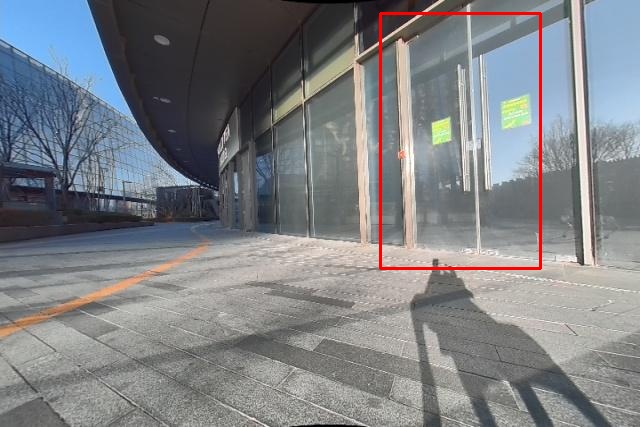}
   \includegraphics[width=0.32\textwidth]{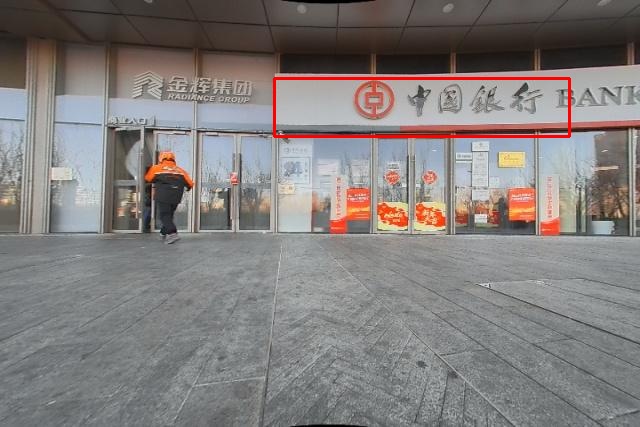}
   \caption{\textbf{Visualization of the latent intention inference module.} Best viewed in color and zoomed in for more details.}
   \label{Fig.5}
\end{figure}

\subsection{Comparisons with Existing Methods}
As the first solution to the out-to-in prior-free instruction-driven embodied navigation task, we found no existing embodied navigation methods reporting evaluation metrics or experimental protocols under this setting. To enable a fair comparison, we reproduce several very recent SOTA method NoMaD, Citywalker and OmniNav, retrain them on our proposed BridgeNav dataset, and uniformly adopt Qwen2.5-VL-3B as the foundation model for vision-language understanding across all baselines.

As shown in Table \ref{tab:table2}, the superiority of our proposed method in navigation success is validated by its significant gains on Success Rate (SR) metrics. Compared to the recent state-of-the-art method OmniNav, our approach improves navigation success rates by \textbf{80.1\%}, \textbf{49.2\%}, and \textbf{23.7\%} under endpoint distance thresholds of 0.1 m, 0.2 m, and 0.3 m, respectively. The improvement on SR at 0.1 m is especially pronounced, indicating that our method excels at precisely reaching the target location. This advantage stems from the latent intention guidance in our framework. As previously described, the latent intention inference module enables the agent to focus on different visual cues at distinct navigation stages, guiding it from coarse-grained outdoor scenes to fine-grained target entrances. The effectiveness of this module is further demonstrated through qualitative analysis in Section \ref{sec5.3}. In terms of navigation efficiency, our method achieves an average improvement of \textbf{31.0\%} over OmniNav, with gains as high as \textbf{66.7\%} in certain environments. This result shows that our approach maintains consistent goal-oriented behavior during navigation, significantly reducing unnecessary detours or redundant turns.


\subsection{Qualitative Results}
\label{sec5.3}
As shown in Fig. \ref{fig4}, we visualize the results of our proposed method deployed on a real-world embodied agent. All navigation instructions are short commands, such as “Go to the nearby cafe,” “I want to deposit money,” or “Go to MOLLY TEA.” During navigation, the agent relies solely on single-frame egocentric (front-facing) images and does not use any additional information such as GPS or pre-built maps. As illustrated in the figure, the agent successfully enters the target location through its entrance/exit in all scenarios. Notably, from the agent’s first-person perspective, multiple candidate locations are often visible simultaneously; correctly identifying the intended destination specified by the instruction among these candidates poses a significant challenge. Another key difficulty lies in locating fine-grained entrance/exit structures within open, unstructured environments. Due to space limitations, we provide additional visualizations in the supplementary material.

\subsection{Ablation Study and Discussion}
\label{sec5.4}
As shown in Table \ref{tab:table3}, we design several ablation variants to further evaluate the effectiveness of each component in our proposed method. First, we remove the initial learnable tokens and instead directly feed the hidden representations from Qwen2.5-VL into an MLP for cross-modal mapping to predict navigation trajectories. In this variant, performance degrades significantly, nearly failing to complete the task altogether. We attribute this to the substantial distributional gap between features derived from vision-language understanding and those required for trajectory prediction; direct modality mapping is thus ill-suited for this task. By introducing initial learnable tokens as task-specific guidance signals, we effectively decouple “understanding” from “decision-making.” These tokens enable the model to adaptively extract, via cross-attention, the most relevant vision-language content for the current navigation step. Second, we investigate the contribution of the intention inference module. Fig. \ref{Fig.5} visualizes intermediate outputs of this module from the first-person perspective of the embodied agent during real-world deployment. In the left panel, we observe that the intention inference module enables the agent to correctly distinguish the target location specified by the instruction among multiple candidate venues. Meanwhile, the middle panel shows that the inferred intention bounding box dynamically adjusts as the agent approaches the goal, signaling that the primary subtask at this stage is to locate the entrance or exit. The right panel illustrates a challenging case: the agent’s view is directly facing a distractor venue, while the instruction is “I want to deposit some money.” Here, the model must not only comprehend the semantic association between “depositing money” and “bank,” but also identify the correct target located in the upper-right corner among several visually similar candidates. Despite this difficulty, as evidenced by the right panel and the fourth column of the first row in Fig. \ref{fig4}, the agent successfully completes the task. Both Table \ref{tab:table3} and our extensive visualizations clearly demonstrate the critical role of this module. Finally, we examine the impact of the optical-flow-guided dynamic perception module. We design a third variant in which we retain the dynamic perception tokens in the encoder but remove the decoding supervision signal during training. As shown in Table \ref{tab:table3}, this ablation confirms that the performance gain stems from the model’s enhanced dynamic perception capability rather than merely from increased parameter count.

\section{Conclusion}
In this work, we define a new task: out-to-in prior-free instruction-driven embodied navigation. To address this task, we propose BridgeNav, a novel vision-centric embodied navigation framework driven primarily by image prompts. Moreover, we introduce a large-scale dataset comprising 2M instruction-trajectory pairs, innovatively constructed using trajectory-conditioned video generation techniques for data synthesis. Extensive experimental results demonstrate the effectiveness and efficiency of our proposed method. We hope this work contributes novel insights to the field of embodied navigation, both methodologically and in terms of dataset resources.
\section*{Impact Statement}
This paper presents work whose goal is to advance the field of Machine
Learning. There are many potential societal consequences of our work, none
which we feel must be specifically highlighted here.

\bibliography{technical_report}
\bibliographystyle{technical_report}

\newpage
\appendix
\onecolumn
\section*{\Large Appendix}
\section{Overview}
This appendix presents more details not included in the main paper due to page limitation.
\begin{itemize}
\item Related work in Sec. B.
\item Dataset details in Sec. C.
\item Experimental details in Sec. D.
\item Qualitative results in Sec. E.
\item Limitations and Future work in Sec. F.
\end{itemize}
\section{Related Work}
\paragraph{Sim-to-Real VLN.}
With the advances in embodied navigation simulators \cite{Matterport3D, Habitat, Ai2-thor, ProcTHOR}, researchers can obtain standardized viewpoint graphs and panoramic observations in controlled indoor environments \cite{VLN-CE}, which has led to the emergence of VLN benchmarks such as R2R \cite{R2R} and RxR \cite{RxR} for systematically evaluating an agent’s navigation capability under language instructions. Existing vision and language navigation methods in simulation can be broadly categorized from multiple perspectives \cite{flexvln}. Some studies leverage Transformer architectures and large-scale pretraining \cite{SkillVln, Navcot, scaleVLN} to improve vision and language alignment and generalization to unseen environments. 
Beyond these, reinforcement learning \cite{BEVBert, RL-VLN, ActiveVLN} is often adopted to optimize long-horizon decision-making. However, when deployed on real-world robotic agents \cite{urbannav}, these methods often suffer from substantial performance degradation \cite{nav-real-world}.

\paragraph{Instruction-Guided Goal Navigation.}
Prior research on long-horizon vision and language navigation \cite{R2R, Towards, languageANDvisual} has largely followed a paradigm with dense instruction-based guidance \cite{episodic, Vln-bert, History}, essentially treating navigation as a deterministic mapping from language to trajectories \cite{counterfactual, Vln-bert}. While formulations provide explicit procedural structure, they also confine agents to passive execution and make them heavily dependent on step-by-step textual cues \cite{Adapt, Diagnosing}. However, last mile target-seeking scenarios expose agents to extremely sparse semantic inputs in the real world \cite{VLN-CE, Vln-bert, Persistent}. Rather than using instructions that provide dense procedural guidance \cite{gadre2023cows, ViNT, Nomad, huang2022visual}, our proposed task starts from lightweight directives at the goal level. 

\section{Dataset Details}
\label{Dataset Details}
\begin{figure*}[h]
  \centering
  \includegraphics[width=1.0\linewidth]{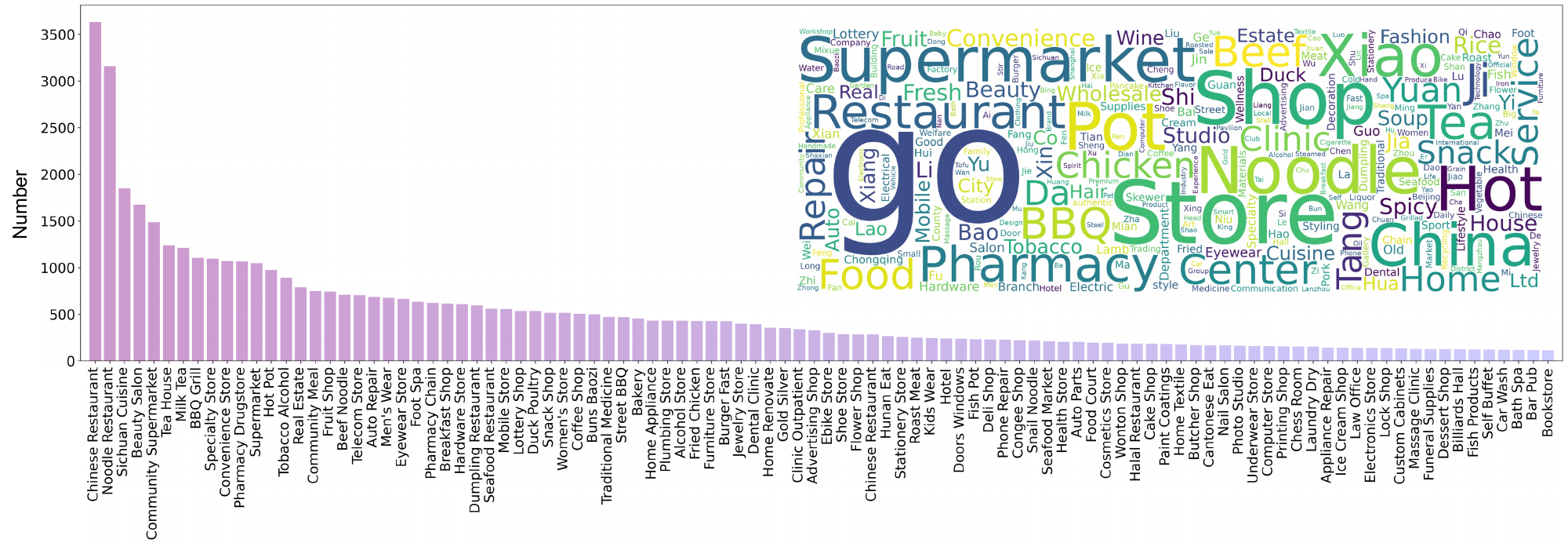}
  \caption{\textbf{Visualization of the BridgeNav Dataset and instruction distribution.}
(1) Top-100 category statistics of BridgeNav Dataset city street-scene samples, shown as a grouped bar chart across different scenario types.
(2) Word cloud of high-frequency tokens in instructions, highlighting the most common concepts and actions that the system is required to understand and execute.}
  \label{Fig.word_cloud}
\end{figure*}

\paragraph{Dataset Statistics and Analysis.}
We visualize the frequency of the top-100 navigation goal categories and the instruction word cloud in Fig. \ref{Fig.word_cloud}. The results confirm that our Dataset encompasses a vast semantic spectrum of urban landmarks, ranging from common commercial storefronts to specialized industrial sites. This diverse distribution enables the training of a generalized navigation policy capable of grounding a wide variety of linguistic targets in complex real-world environments.

\paragraph{Local Occupancy Grid Mapping.}
\label{Local Occupancy Grid Mapping.}
Provision of sufficiently fine-grained spatial descriptions and the assurance of navigation without collisions are achieved through the construction of a local occupancy grid for each street view image. Firstly, the extraction of dense depth information from raw images utilizes the MoGe-v2 \cite{Moge} depth estimation model, followed by the transformation of this data into 3D point clouds. Isolation of navigable areas within complex unstructured spaces involves the identification of ground planes and obstacles by calculating surface normals for each point in the cloud. Specifically, points with normals oriented vertically upward are categorized as ground or navigable regions, whereas those with horizontal or inconsistent orientations represent obstacles such as walls, utility poles, or pedestrians. Building upon this segmentation, the projection of the point cloud onto a 2D top-down plane generates an agent-centric local occupancy grid. The results are represented in a $50 \times 50$ occupancy grid in the perception range
$\left[-2.5\,\mathrm{m} \sim 2.5\,\mathrm{m}, 0\,\mathrm{m} \sim 5\,\mathrm{m}\right]$
along the X–Z (horizontal) plane. Concurrently, mapping of the identified navigation target onto grid coordinate indices is performed by integrating visual cues with depth information. For targets exceeding the perception range, an anchoring strategy at the nearest edge along the Z-axis ensures the validity of the local navigation objective.
Discretization of the environment into free and occupied states via this grid provides the essential geometric constraints for subsequent path planning in complex terminal urban scenes.

\begin{figure}[tb]
  \centering
  \includegraphics[width=\linewidth,height=0.22\textheight,keepaspectratio]{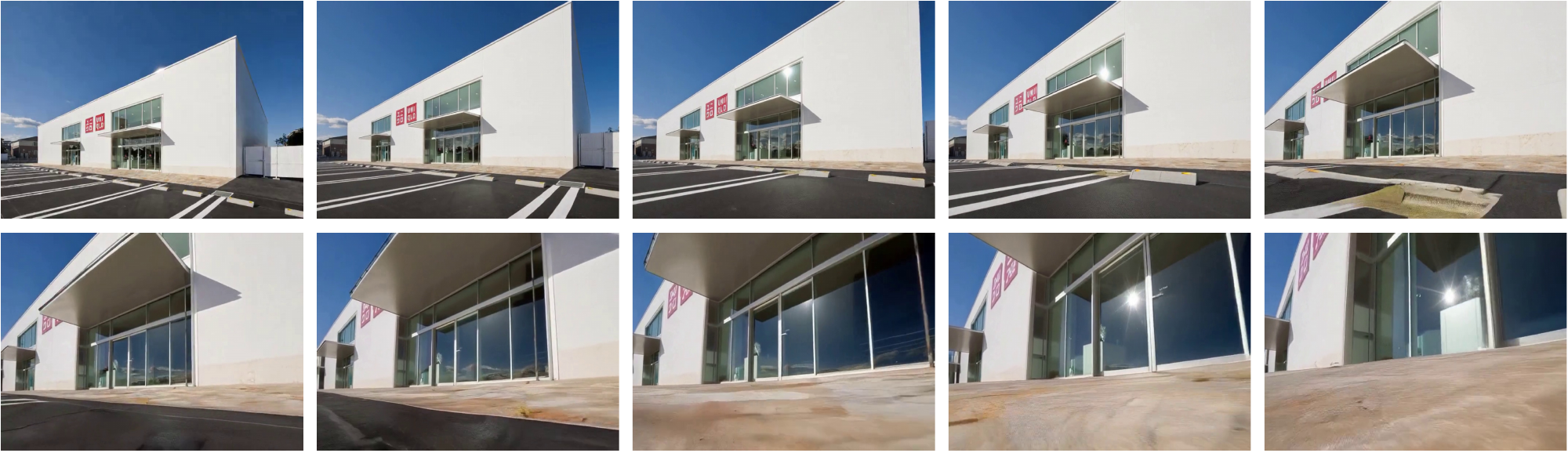}\par\smallskip
  \includegraphics[width=\linewidth,height=0.22\textheight,keepaspectratio]{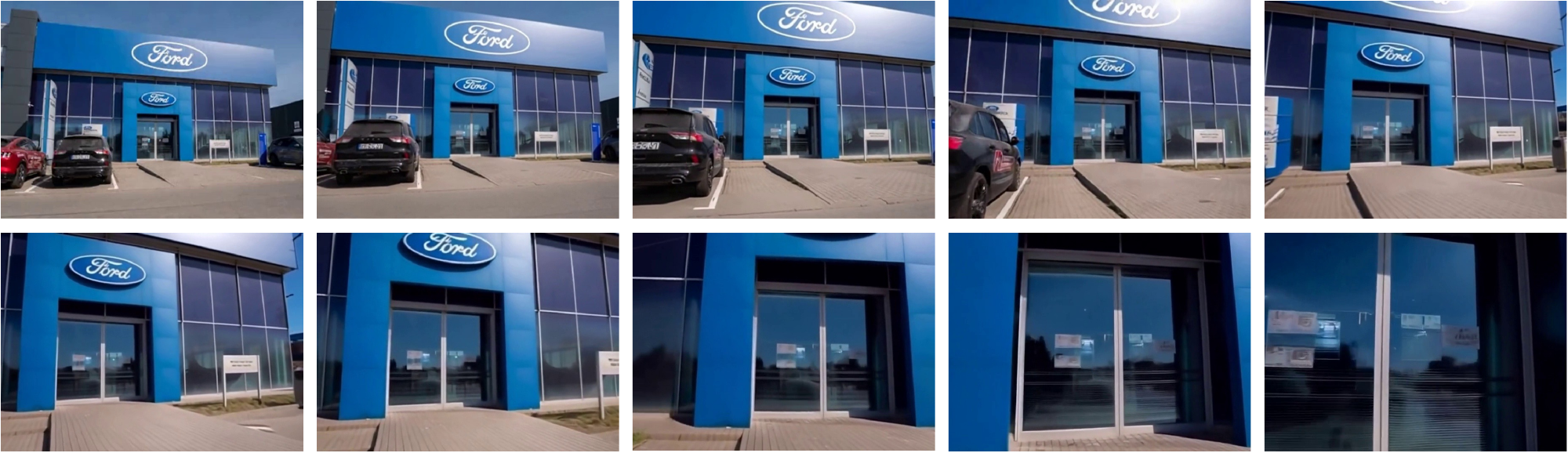}\par\smallskip
  \includegraphics[width=\linewidth,height=0.22\textheight,keepaspectratio]{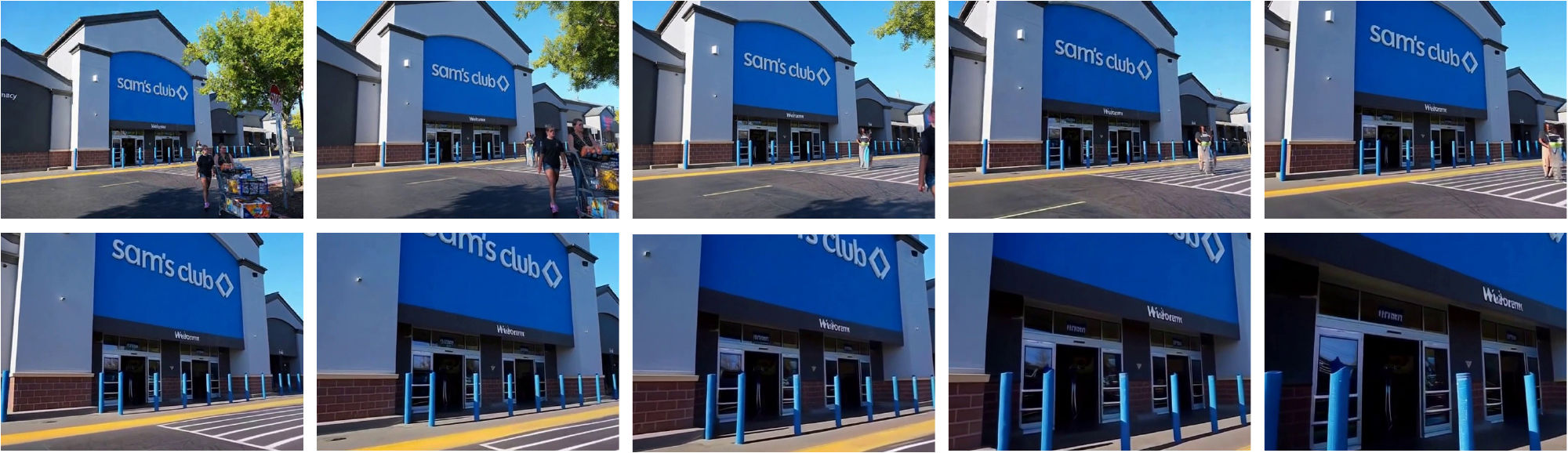}\par\smallskip
  \includegraphics[width=\linewidth,height=0.22\textheight,keepaspectratio]{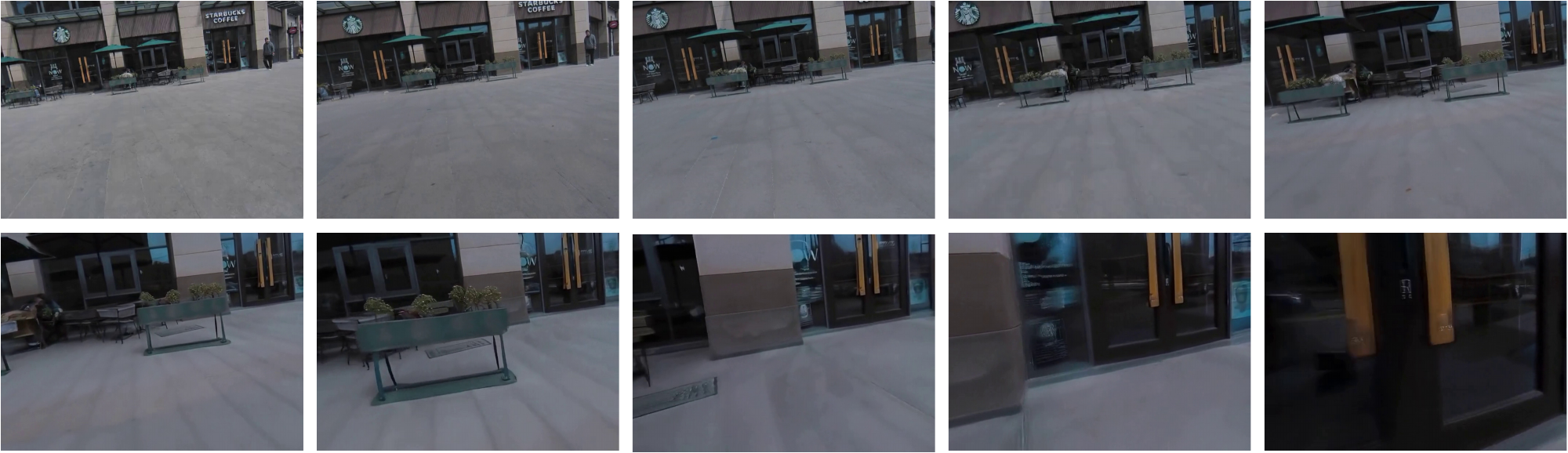}
  \caption{\textbf{Example samples from our proposed dataset.} Best viewed in color and zoomed in for more details.}
  \label{fig:show_case_4x1}
\end{figure}

\section{Experimental Details}
\label{Appendix-Experimental Details}
Our BridgeNav framework is implemented in PyTorch and trained using the Adam optimizer under a cosine learning rate schedule. The initial learning rate is set to $1\times10^{-5}$. In the first training stage, we train for approximately 5 hours on 8 NVIDIA H20 GPUs with a total batch size of 64. In the second stage, training proceeds for about 48 hours on 128 NVIDIA H20 GPUs with a total batch size of 256. During real-world deployment, our method runs at over 5 Hz on a single NVIDIA A10 GPU.

\section{Qualitative Results}
\begin{figure*}[tbp]
  \centering
    \includegraphics[width=1\linewidth]{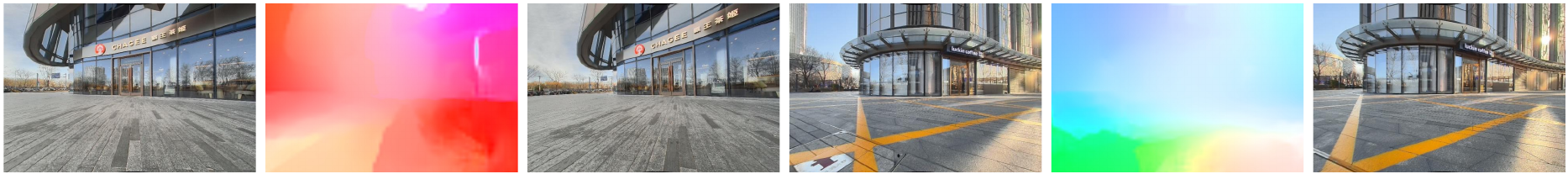}
  \caption{\textbf{Visualization of optical flow.} Best viewed in color and zoomed in for more details.}
  \label{optical-flow}
\end{figure*}

As shown in Fig. \ref{optical-flow}, we visualize the optical flow between two consecutive frames. The regions with the highest optical flow magnitude accurately reflect the direction of motion, which we leverage to guide the model in establishing the mapping between navigation trajectories and observed images. 

\begin{figure}[tbp]
  \centering
  \includegraphics[width=\linewidth,height=0.22\textheight,keepaspectratio]{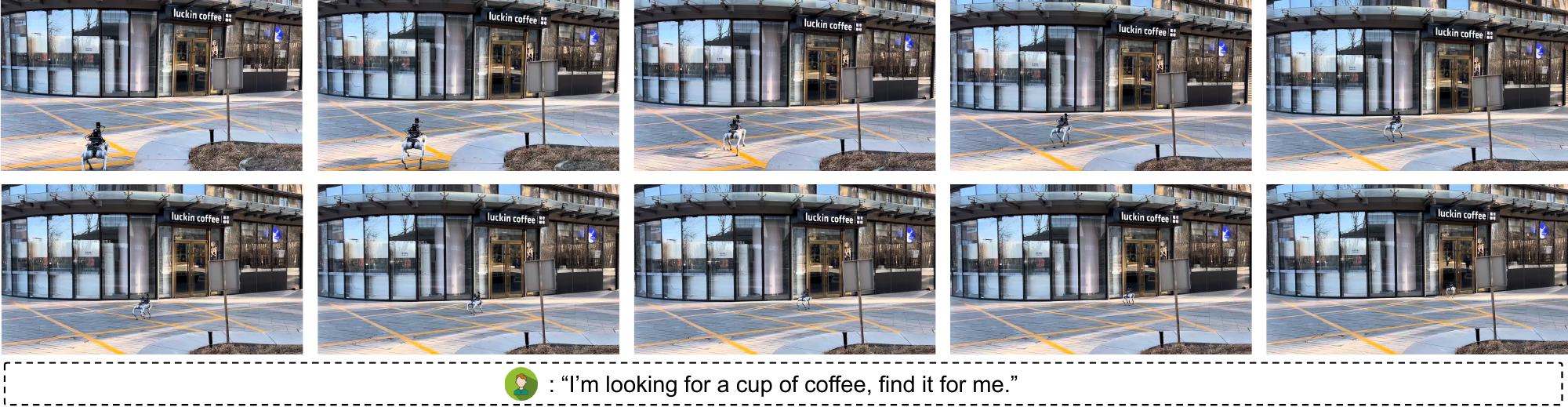}\par\smallskip
  \includegraphics[width=\linewidth,height=0.22\textheight,keepaspectratio]{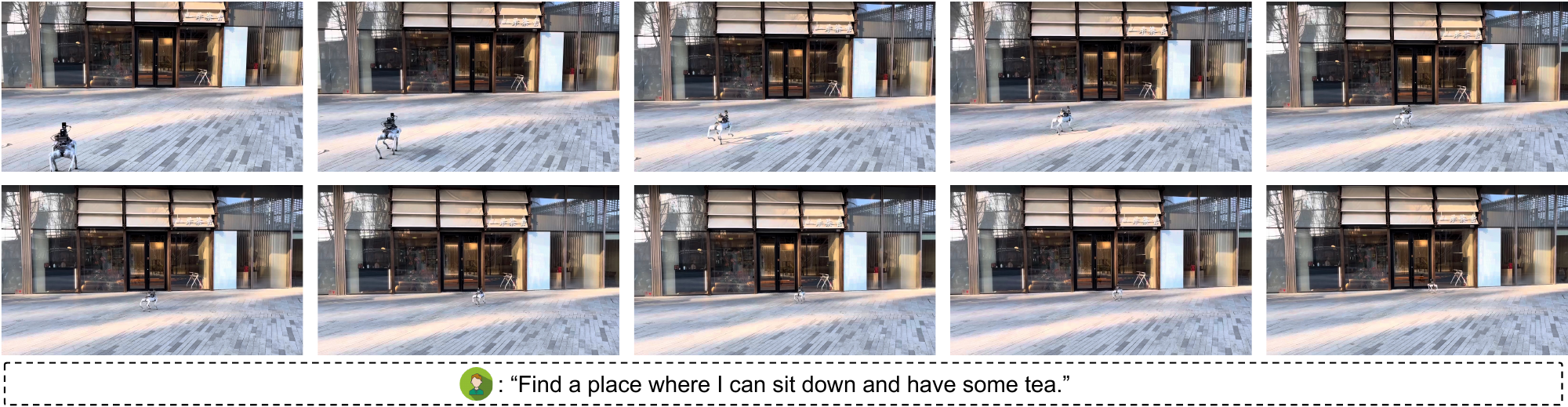}\par\smallskip
  \includegraphics[width=\linewidth,height=0.22\textheight,keepaspectratio]{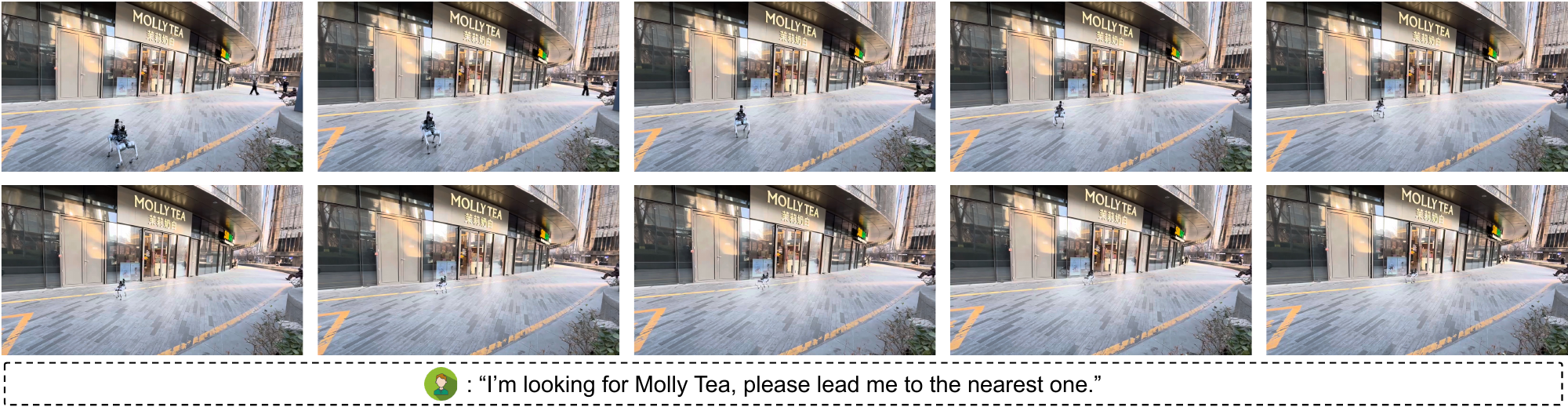}\par\smallskip
  \includegraphics[width=\linewidth,height=0.22\textheight,keepaspectratio]{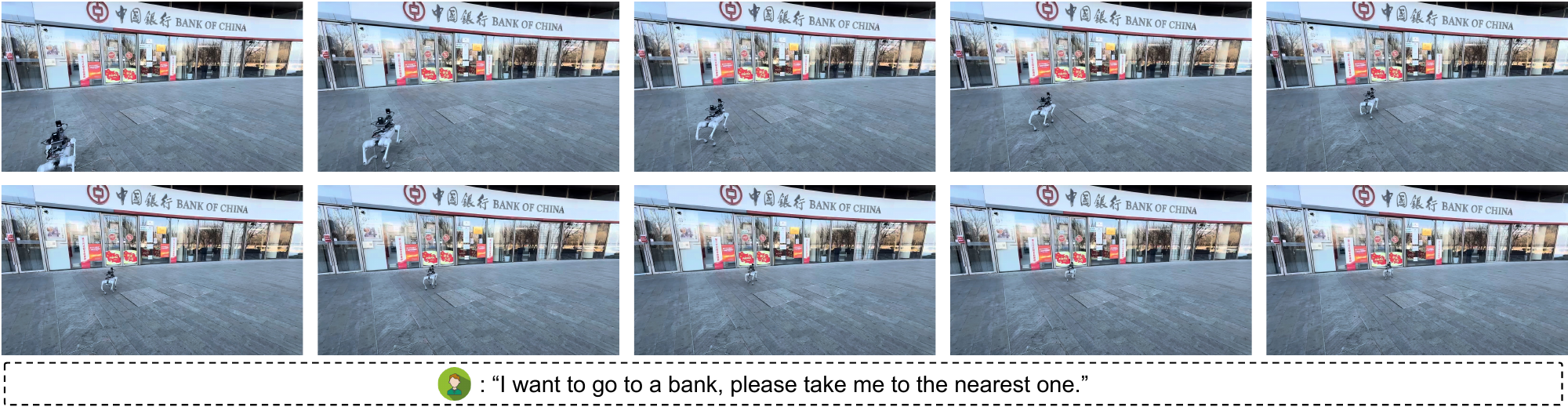}
  \caption{\textbf{Example samples from our real-world deployment tests.} Best viewed in color and zoomed in for more details.}
  \label{fig:show_demo_4x1}
\end{figure}

As shown in Fig. \ref{fig:show_demo_4x1}, we present several scenarios from our real-world deployment tests. Note that, due to the limitations of the camera viewpoint used for recording, only the target location mentioned in the instruction is visible in these images. In reality, however, the embodied quadruped robot’s first-person observations (see Fig. \ref{Fig.5}) often contain multiple candidate locations within its field of view. The deployment results demonstrate that our proposed framework can not only accurately identify the correct target among these candidates but also precisely navigate through the entrance.

\section{Limitations and Future Work}
However, BridgeNav still has some limitations: (1) During real-world deployment, we observed that BridgeNav performs poorly when the input images suffer from distortion or are of excessively low resolution. (2) Our current implementation and evaluation are limited to quadrupedal robots and have not yet been extended to humanoid robots. Therefore, to enable delivery robots to truly integrate into our daily lives, the robustness and generalizability of the method require further improvement. We leave them for future work.

\end{document}